\documentclass{templates/llncs} 

\usepackage{IEEEtrantools}

\usepackage{etoolbox}
\newcommand{\repthanks}[1]{\textsuperscript{\ref{#1}}}
\makeatletter
\patchcmd{\maketitle}
{\def\thanks}
{\let\repthanks\repthanksunskip\def\thanks}
{}{}
\patchcmd{\@maketitle}
{\def\thanks}
{\let\repthanks\@gobble\def\thanks}
{}{}
\newcommand\repthanksunskip[1]{\unskip{}}
\makeatother
\usepackage{physics}
\usepackage{graphicx}
\usepackage{graphics}
\usepackage{gensymb}
\usepackage{booktabs}
\usepackage{siunitx}
\usepackage{floatrow}

\usepackage{IEEEtrantools}
\bstctlcite{IEEEexample:BSTcontrol}

\usepackage{multirow}
\newfloatcommand{capbtabbox}{table}[][\FBwidth]

\usepackage{xcolor}
\usepackage[colorlinks = true,
            linkcolor = blue,
            urlcolor  = blue,
            citecolor = blue,
            anchorcolor = blue]{hyperref}

\newcommand{\MYhref}[3][blue]{\href{#2}{\color{#1}{#3}}}%

\begin{document}

\bstctlcite{IEEEexample:BSTcontrol}

\title{nnU-Net: Self-adapting Framework \\for U-Net-Based Medical Image Segmentation}



\author{
Fabian Isensee \and
Jens Petersen \and
Andre Klein \and
David Zimmerer \and
Paul F. Jaeger \and
Simon Kohl \and
Jakob Wasserthal \and
Gregor K\"ohler \and
Tobias Norajitra \and
Sebastian Wirkert \and
Klaus H. Maier-Hein}
\authorrunning{Fabian Isensee et al.} 
%
\tocauthor{**************, **************, **************}
\institute{Division of Medical Image Computing, German Cancer Research Center (DKFZ),\\ Heidelberg, Germany
}

\maketitle   

\begin{abstract}

The U-Net was presented in 2015. With its straight-forward and successful architecture it quickly evolved to a commonly used benchmark in medical image segmentation. The adaptation of the U-Net to novel problems, however, comprises several degrees of freedom regarding the exact architecture, pre-processing, training and inference. These choices are not independent of each other and substantially impact the overall performance. The present paper introduces the nnU-Net ("no-new-Net"), which refers to a robust and self-adapting framework on the basis of 2D and 3D vanilla U-Nets. We argue the strong case for taking away superfluous bells and whistles of many proposed network designs and instead focus on the remaining aspects that make out the performance and generalizability of a method. We evaluate the nnU-Net in the context of the Medical Segmentation Decathlon challenge, which measures segmentation performance in ten disciplines comprising distinct entities, image modalities, image geometries and dataset sizes, with no manual adjustments between datasets allowed. At the time of manuscript submission, nnU-Net achieves the highest mean dice scores across all classes and seven phase 1 tasks (except class 1 in BrainTumour) in the online leaderboard of the challenge.

%
\end{abstract}
\section{Introduction}
    Medical Image Segmentation is currently dominated by deep convolutional neural networks (CNNs). However, each segmentation benchmark seems to require specialized architectures and training scheme modifications to achieve competitive performance \cite{isensee2017brain,li2017h,roy2018concurrent,oktay2018attention,jegou2017one}. This results in huge amounts of publications in the field that, alongside often limited validation on only few or even just a single dataset, make it increasingly difficult for researchers to identify methods that live up to their promised superiority beyond the limited scenarios they are demonstrated on. The Medical Segmentation Decathlon is intended to specifically address this issue: participants in this challenge are asked to create a segmentation algorithm that generalizes across 10 datasets corresponding to different entities of the human body. These algorithms may dynamically adapt to the specifics of a particular dataset, but are only allowed to do so in a fully automatic manner. The challenge is split into two successive phases: 1) a development phase in which participants are given access to 7 datasets to optimize their approach on and, using their final and thus frozen method, must submit segmentations for the corresponding 7 held-out test sets. 2) a second phase to evaluate the same exact method on 3 previously undisclosed datasets.
    
    We hypothesize that some of the architectural modifications presented recently are in part overfitted to specific problems or could suffer from imperfect validation that results from sub-optimal reimplementations of the state-of-the-art. Using the U-Net as a benchmark on an in-house dataset, for example, requires the adaptation of the method to the novel problem. This spans several degrees of freedom. Even though the architecture itself is quite straight-forward, and even though the method is quite commonly used as a benchmark, we believe that the remaining interdependent choices regarding the exact architecture, pre-processing, training, inference and post-processing quite often cause the U-Net to underperform when used as a benchmark. Additionally, architectural tweaks that are intended to improve the performance of a network can rather easily be demonstrated to work if the network is not yet fully optimized for the task at hand, allowing for plenty of headroom for the tweak to improve results. In our own preliminary experiments, these tweaks however were unable to improve segmentation results in fully optimized networks and thus most likely unable to advance the state of the art. This leads us to believe that the influence of non-architectural aspects in segmentation methods is much more impactful, but at the same time also severely underestimated. 
    
    In this paper, we present the nnU-Net ("no-new-Net") framework. It resides on a set of three comparatively simple U-Net models that contain only minor modifications to the original U-Net \cite{ronneberger2015u}. We omit recently proposed extensions such as for example the use of residual connections \cite{he2016identity,milletari2016v}, dense connections \cite{jegou2017one} or attention mechanisms \cite{oktay2018attention}. The nnU-Net automatically adapts its architectures to the given image geometry. More importantly though, the nnU-Net framework thoroughly defines all the other steps around them. These are steps where much of the nets' performance can be gained or respectively lost: preprocessing (e.g. resampling and normalization), training (e.g. loss, optimizer setting and data augmentation), inference (e.g. patch-based strategy and ensembling across test-time augmentations and models) and a potential post-processing (e.g. enforcing single connected components if applicable).

\section{Methods}
    \label{methods}
    \subsection{Network architectures}
    \label{networkarchitecture}
    
    Medical images commonly encompass a third dimension, which is why we consider a pool of basic U-Net architectures consisting of a 2D U-Net, a 3D U-Net and a U-Net Cascade. While the 2D and 3D U-Nets generate segmentations at full resolution, the cascade first generates low resolution segmentations and subsequently refines them. Our architectural modifications as compared to the U-Net's original formulation are close to negligible and instead we focus our efforts on designing an automatic training pipeline for these models.
    
    The U-Net \cite{ronneberger2015u} is a successful encoder-decoder network that has received a lot of attention in the recent years. Its encoder part works similarly to a traditional classification CNN in that it successively aggregates semantic information at the expense of reduced spatial information. Since in segmentation, both semantic as well as spatial information are crucial for the success of a network, the missing spatial information must somehow be recovered. The U-Net does this through the decoder, which receives semantic information from the bottom of the 'U' and recombines it with higher resolution feature maps obtained directly from the encoder through skip connections. Unlike other segmentation networks, such as FCN \cite{long2015fully} and previous iterations of DeepLab \cite{chen2018deeplab} this allows the U-Net to segment fine structures particularly well.
    
    Just like the original U-Net, we use two plain convolutional layers between poolings in the encoder and transposed convolution operations in the decoder. We deviate from the original architecture in that we replace ReLU activation functions with leaky ReLUs (neg. slope $1e^{-2}$) and use instance normalization~\cite{ulyanov2016instance} instead of the more popular batch normalization \cite{ioffe2015batch}. 

    \subsubsection{2D U-Net}
    Intuitively, using a 2D U-Net in the context of 3D medical image segmentation appears to be suboptimal because valuable information along the z-axis cannot be aggregated and taken into consideration. However, there is evidence \cite{isensee2017automatic} that conventional 3D segmentation methods deteriorate in performance if the dataset is anisotropic  (cf. Prostate dataset of the Decathlon challenge).
    
    \subsubsection{3D U-Net}
    A 3D U-Net seems like the appropriate method of choice for 3D image data. In an ideal world, we would train such an architecture on the entire patient's image. In reality however, we are limited by the amount of available GPU memory which allows us to train this architecture only on image patches. While this is not a problem for datasets comprised of smaller images (in terms of number of voxels per patient) such as the Brain Tumour, Hippocampus and Prostate datasets of this challenge, patch-based training, as dictated by datasets with large images such as Liver, may impede training. This is due to the limited field of view of the architecture which thus cannot collect sufficient contextual information to e.g. correctly distinguish parts of a liver from parts of other organs.
    
    \begin{figure}
        \centering
        \includegraphics[width=\textwidth]{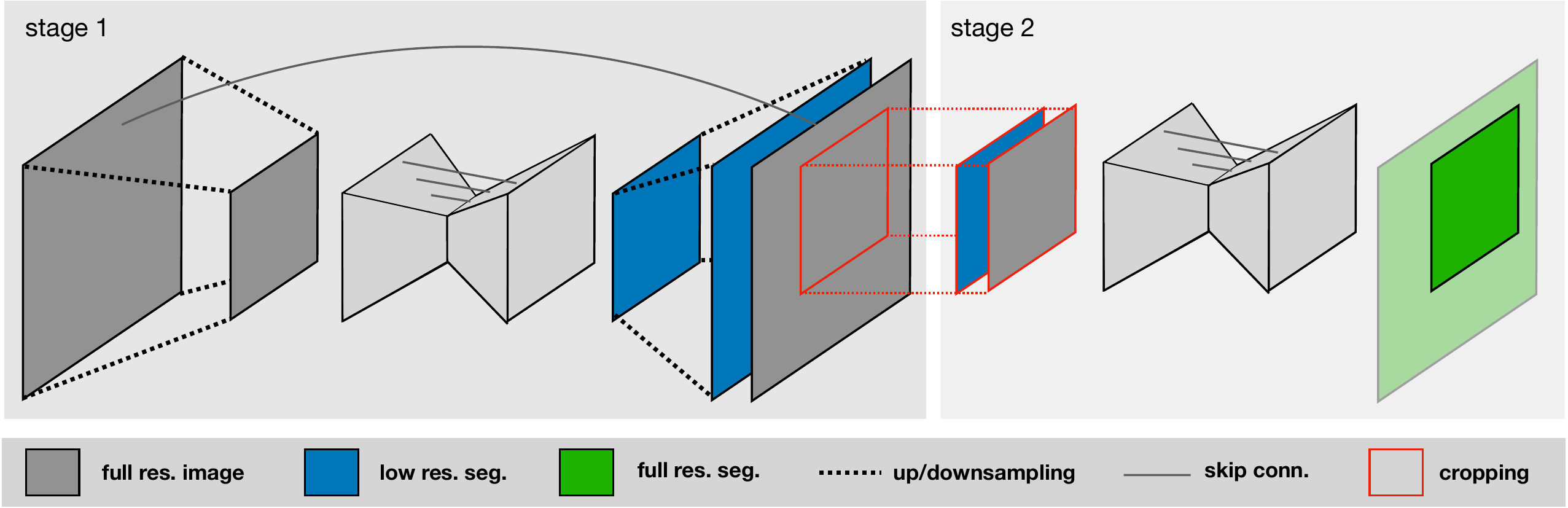}
        \caption{U-Net Cascade (on applicable datasets only). Stage 1 (left): a 3D U-Net processes downsampled data, the resulting segmentation maps are upsampled to the original resolution. Stage 2 (right): these segmentations are concatenated as one-hot encodings to the full resolution data and refined by a second 3D U-Net.}
        \label{fig:cascade}
    \end{figure}

    \subsubsection{U-Net Cascade}
    To address this practical shortcoming of a 3D U-Net on datasets with large image sizes, we additionally propose a cascaded model. Therefore, a 3D U-Net is first trained on downsampled images (stage 1). The segmentation results of this U-Net are then upsampled to the original voxel spacing and passed as additional (one hot encoded) input channels to a second 3D U-Net, which is trained on patches at full resolution (stage 2). See Figure \ref{fig:cascade}.
    
    \subsubsection{Dynamic adaptation of network topologies}
    Due to the large differences in image size (median shape $482 \times 512 \times 512$ for Liver vs. $36 \times 50 \times 35$ for Hippocampus) the input patch size and number of pooling operations per axis (and thus implicitly the number of convolutional layers) must be automatically adapted for each dataset to allow for adequate aggregation of spatial information. Apart from adapting to the image geometries, there are technical constraints like the available memory to account for. Our guiding principle in this respect is to dynamically trade off the batch-size versus the network capacity, presented in detail below:
    
    We start out with network configurations that we know to be working with our hardware setup. For the 2D U-Net this configuration is an input patch size of $256 \times 256$, a batch size of 42 and 30 feature maps in the highest layers (number of feature maps doubles with each downsampling). We automatically adapt these parameters to the median plane size of each dataset (where we use the plane with the lowest in-plane spacing, corresponding to the highest resolution), so that the network effectively trains on entire slices. We configure the networks to pool along each axis until the feature map size for that axis is smaller than 8 (but not more than a maximum of 6 pooling operations). 
    Just like the 2D U-Net, our 3D U-Net uses 30 feature maps at the highest resolution layers. Here we start with a base configuration of input patch size $128 \times 128 \times 128$, and a batch size of 2. Due to memory constraints, we do not increase the input patch volume beyond $128^3$ voxels, but instead match the aspect ratio of the input patch size to that of the median size of the dataset in voxels. If the median shape of the dataset is smaller than $128^3$ then we use the median shape as input patch size and increase the batch size (so that the total number of voxels processed is the same as with $128 \times 128 \times 128$ and a batch size of 2). Just like for the 2D U-Net we pool (for a maximum of 5 times) along each axis until the feature maps have size 8.
    
    For any network we limit the total number of voxels processed per optimizer step (defined as the input patch volume times the batch size) to a maximum of 5\% of the dataset. For cases in excess, we reduce the batch size (with a lower-bound of 2). 

    All network topologies generated for the phase 1 datasets are presented in table \ref{segmentation_architectures}.
    
    \begin{table}[t]
    \label{segmentation_architectures}
    \scriptsize
    \begin{tabular}{ll|ccc}
    & & 2D U-Net & 3D U-Net & 3D U-Net lowres \\
    \hline
    \multirow{4}{*}{BrainTumour} & median patient shape & 169x138 & 138x169x138 & - \\
     & input patch size & 192x160 & 128x128x128 & - \\
     & batch size & 89 & 2 & - \\
     & num pool per axis & 5, 5 & 5, 5, 5 & - \\
     \hline
    \multirow{4}{*}{Heart} & median patient shape & 320x232 & 115x320x232 & 58x160x116 \\
     & input patch size & 320x256 & 80x192x128 & 64x160x128 \\
     & batch size & 33 & 2 & 2 \\
     & num pool per axis & 6, 6 & 4, 5, 5 & 4, 5, 5 \\
    \hline
    \multirow{4}{*}{Liver} & median patient shape & 512x512 & 482x512x512 & 121x128x128 \\
     & input patch size & 512x512 & 128x128x128 & 128x128x128 \\
     & batch size & 10 & 2 & 2 \\
     & num pool per axis & 6, 6 & 5, 5, 5 & 5, 5, 5 \\
    \hline
    \multirow{4}{*}{Hippocampus} & median patient shape & 50x35 & 36x50x35 & - \\
     & input patch size & 56x40 & 40x56x40 & - \\
     & batch size & 366 & 9 & - \\
     & num pool per axis & 3, 3 & 3, 3, 3 & - \\
    \hline
    \multirow{4}{*}{Prostate} & median patient shape & 320x319 & 20x320x319 & - \\
     & input patch size & 320x320 & 20x192x192 & - \\
     & batch size & 26 & 4 & - \\
     & num pool per axis & 6, 6 & 2, 5, 5 & - \\
    \hline
    \multirow{4}{*}{Lung} & median patient shape & 512x512 & 252x512x512 & 126x256x256 \\
     & input patch size & 512x512 & 112x128x128 & 112x128x128 \\
     & batch size & 10 & 2 & 2 \\
     & num pool per axis & 6, 6 & 4, 5, 5 & 4, 5, 5 \\
    \hline
    \multirow{4}{*}{Pancreas} & median patient shape & 512x512 & 96x512x512 & 96x256x256 \\
     & input patch size & 512x512 & 96x160x128 & 96x160x128 \\
     & batch size & 10 & 2 & 2 \\
     & num pool per axis & 6, 6 & 4, 5, 5 & 4, 5, 5\\
     \hline
    \end{tabular}
    \caption{Network topologies as automatically generated for the seven phase 1 tasks of the Medical Segmentation Decathlon challenge. 3D U-Net lowres refers to the first stage of the U-Net Cascade. The configuration of the second stage of the U-Net Cascade is identical to the 3D U-Net.}
    \end{table}
    \subsection{Preprocessing}
    The preprocessing is part of the fully automated segmentation pipeline that our method consists of and, as such, the steps presented below are carried out without any user intervention. 
    
    \subsubsection{Cropping}
    All data is cropped to the region of nonzero values. This has no effect on most datasets such as liver CT, but will reduce the size (and therefore the computational burden) of skull stripped brain MRI. 
    
    \subsubsection{Resampling}
    \label{resampling}
    CNNs do not natively understand voxel spacings. In medical images, it is common for different scanners or different acquisition protocols to result in datasets with heterogeneous voxel spacings. To enable our networks to properly learn spatial semantics, all patients are resampled to the median voxel spacing of their respective dataset, where third order spline interpolation is used for image data and nearest neighbor interpolation for the corresponding segmentation mask.
    
    Necessity for the U-Net Cascade is determined by the following heuristics: If the median shape of the resampled data has more than 4 times the voxels that can be processed as input patch by the 3D U-Net (with a batch size of 2), it qualifies for the U-Net Cascade and this dataset is additionally resampled to a lower resolution. This is done by increasing the voxel spacing (decrease resolution) by a factor of 2 until the above mentioned criterion is met. If the dataset is anisotropic, the higher resolution axes are first downsampled until they match the low resolution axis/axes and only then all axes are downsampled simultaneously. The following datasets of phase 1 fall within the set of described heuristics and hence trigger usage of the U-Net Cascade: Heart, Liver, Lung, and Pancreas.
    
    \subsubsection{Normalization}
    Because the intensity scale of CT scans is absolute, all CT images are automatically normalized based on statistics of the entire respective dataset: If the modality description in a dataset's corresponding json desccriptor file indicates `ct', all intensity values occurring within the segmentation masks of the training dataset are collected and the entire dataset is normalized by clipping to the [0.5, 99.5] percentiles of these intensity values, followed by a z-score normalization based on the mean and standard deviation of all collected intensity values. For MRI or other image modalities (i.e. if no `ct' string is found in the modality), simple z-score normalization is applied to the patient individually.
    
    If cropping reduces the average size of patients in a dataset (in voxels) by 1/4 or more the normalization is carried out only within the mask of nonzero elements and all values outside the mask are set to 0.

    \subsection{Training Procedure}
    All models are trained from scratch and evaluated using five-fold cross-validation on the training set. We train our networks with a combination of dice and cross-entropy loss:
    \begin{equation}
        \mathcal{L}_{total} = \mathcal{L}_{dice} + \mathcal{L}_{CE}
    \end{equation}
    
    For 3D U-Nets operating on nearly entire patients (first stage of the U-Net Cascade and 3D U-Net if no cascade is necessary) we compute the dice loss for each sample in the batch and average over the batch. For all other networks we interpret the samples in the batch as a pseudo-volume and compute the dice loss over all voxels in the batch.
    
    The dice loss formulation used here is a multi-class adaptation of the variant proposed in  \cite{drozdzal2016importance}. Based on past experience \cite{isensee2017automatic,isensee2017brain} we favor this formulation over other variants \cite{milletari2016v,sudre2017generalised}. The dice loss is implemented as follows:
    \begin{equation}
    \mathcal{L}_\mathrm{dc} = - \frac{2}{|K|} \sum_{k\in K}\frac{\sum_{i\in I} u_i^k v_i^k}{\sum_{i\in I} u_i^k + \sum_{i\in I} v_i^k}
    \end{equation}
    
    where $u$ is the softmax output of the network and $v$ is a one hot encoding of the ground truth segmentation map. Both $u$ and $v$ have shape $I \times K$ with $i \in I$ being the number of pixels in the training patch/batch and $k\in K$ being the classes.
    
    We use the Adam optimizer with an initial learning rate of $3\times10^{-4}$ for all experiments. We define an epoch as the iteration over 250 training batches. During training, we keep an exponential moving average of the validation ($l_{MA}^{v}$) and training ($l_{MA}^{t}$) losses. Whenever $l_{MA}^{t}$ did not improve by at least $5\times 10^{-3}$ within the last 30 epochs, the learning rate was reduced by factor 5. The training was terminated automatically if $l_{MA}^{v}$ did not improve by more than $5\times 10^{-3}$ within the last 60 epochs, but not before the learning rate was smaller than $10^{-6}$.
    
    \subsubsection{Data Augmentation}
    When training large neural networks from limited training data, special care has to be taken to prevent overfitting. We address this problem by utilizing a large variety of data augmentation techniques. The following augmentation techniques were applied on the fly during training: random rotations, random scaling, random elastic deformations, gamma correction augmentation and mirroring. Data augmentation was done with our own in-house framework which is publically available at
    \MYhref{https://github.com/MIC-DKFZ/batchgenerators}{github.com/MIC-DKFZ/batchgenerators}.
    
    We define sets of data augmentation parameters for the 2D and 3D U-Net separately. These parameters are not modified between datasets.
    
    Applying three dimensional data augmentation may be suboptimal if the maximum edge length of the input patch size of a 3D U-Net is more than two times as large as the shortest. For datasets where this criterion applies we use our 2D data augmentation instead and apply it slice-wise for each sample.
    
    The second stage of the U-Net Cascade receives the segmentations of the previous step as additional input channels. To prevent strong co-adaptation we apply random morphological operators (erode, dilate, open, close) and randomly remove connected components of these segmentations.
    
    \subsubsection{Patch Sampling}
    To increase the stability of our network training we enforce that more than a third of the samples in a batch contain at least one randomly chosen foreground class.
    
    \subsection{Inference}
    Due to the patch-based nature of our training, all inference is done patch-based as well. Since network accuracy decreases towards the border of patches, we weigh voxels close to the center higher than those close to the border, when aggregating predictions across patches. Patches are chosen to overlap by patch size / 2 and we further make use of test time data augmentation by mirroring all patches along all valid axes.
    
    Combining the tiled prediction and test time data augmentation result in segmentations where the decision for each voxel is obtained by aggregating up to 64 predictions (in the center of a patient using 3D U-Net). For the test cases we use the five networks obtained from our training set cross-validation as an ensemble to further increase the robustness of our models.
    
    \subsection{Postprocessing}
    A connected component analysis of all ground truth segmentation labels is performed on the training data. If a class lies within a single connected component in all cases, this behaviour is interepreted as a general property of the dataset. Hence, all but the largest connected component for this class are automatically removed on predicted images of the corresponding dataset.
    
    \subsection{Ensembling and Submission}
    To further increase the segmentation performance and robustness all possible combinations of two out of three of our models are ensembled for each dataset. For the final submission, the model (or ensemble) that achieves the highest mean foreground dice score on the training set cross-validation is automatically chosen.

\section{Experiments and Results}
    We optimize our network topologie using five-fold cross-validations on the phase 1 datasets. Our phase 1 cross-validation results as well as the corresponding submitted test set results are summarized in Table \ref{tab:results}. - indicates that the U-Net Cascade was not applicable (i.e. necessary, according to our criteria) to a dataset because it was already fully covered by the input patch size of the 3D U-Net. The model that was used for the final submission is highlighted in bold. Although several test set submissions were allowed by the platform, we believe it to be bad practice to do so. Hence we only submitted once and report the results of this single submission. 
    
    As can be seen in Table \ref{tab:results} our phase 1 cross-validation results are robustly recovered on the held-out test set indicating a desired absence of over-fitting. The only dataset that suffers from a dip in performance on all of its foreground classes is BrainTumour. The data of this phase 1 dataset stems from the BRATS challenge \cite{menze2015multimodal} for which such performance drops between validation and testing are a common sight and attributed to a large shift in the respective data and/or ground-truth distributions.

\begin{table}[h]
\centering
\scalebox{0.92}{
\scriptsize
\begin{tabular}{l|ccc|c|cc|cc|cc|c|cc}
 & \multicolumn{3}{c}{BrainTumour} & \multicolumn{1}{c}{Heart} & \multicolumn{2}{c}{Liver} & \multicolumn{2}{c}{Hippoc.} & \multicolumn{2}{c}{Prostate} & \multicolumn{1}{c}{Lung} & \multicolumn{2}{c}{Pancreas} \\
label & 1 & 2 & 3 & 1 & 1 & 2 & 1 & 2 & 1 & 2 & 1 & 1 & 2 \\
 \hline
\begin{tabular}[c]{@{}l@{}} \\ 2D U-Net \\ \, \end{tabular} & 78.60 & 58.65 & 77.42 & 91.36 & 94.37 & 53.94 & 88.52 & 86.70 & 61.98 & 84.31 & 52.68 & 74.70 & 35.41 \\
\hline
\begin{tabular}[c]{@{}l@{}} \\ 3D U-Net \\ \, \end{tabular} & \textbf{80.71} & \textbf{62.22} & \textbf{79.07} & 92.45 & 94.11 & 61.74 & \textbf{89.87} & \textbf{88.20} & 60.77 & 83.73 & 55.87 & 77.69 & 42.69 \\
\hline
\begin{tabular}[c]{@{}l@{}}3D U-Net \\ stage1 only \\ (U-Net Cascade)\end{tabular} & - & - & - & 90.63 & 94.69 & 47.01 & - & - & - & - & 65.33 & 79.45 & 49.65 \\
\hline
\begin{tabular}[c]{@{}l@{}}3D U-Net \\ (U-Net Cascade)\end{tabular} & - & - & - & 92.40 & 95.38 & 58.49 & - & - & - & - & \textbf{66.85} & \textbf{79.30} & \textbf{52.12} \\
\hline
\begin{tabular}[c]{@{}l@{}}ensemble \\ 2D U-Net+ \\ 3D U-Net\end{tabular} & 80.79 & 61.72 & 79.16 & \textbf{92.70} & 94.30 & 60.24 & 89.78 & 88.09 & \textbf{63.78} & \textbf{85.31} & 55.96 & 78.26 & 40.46 \\
\hline
\begin{tabular}[c]{@{}l@{}}ensemble \\ 2D U-Net+ \\ 3D U-Net \\ (U-Net Cascade)\end{tabular} & - & - & - & 92.64 & 95.31 & 60.09 & - & - & - & - & 61.18 & 78.79 & 45.46 \\
\hline
\begin{tabular}[c]{@{}l@{}}ensemble \\ 3D U-Net+ \\ 3D U-Net \\ (U-Net Cascade)\end{tabular} & - & - & - & 92.63 & \textbf{95.43} & \textbf{61.82} & - & - & - & - & 65.16 & 79.70 & 49.14 \\
\hline
\hline
\begin{tabular}[c]{@{}l@{}} \\ test set \\ \, \end{tabular} & 67.71 & \textbf{47.73} & \textbf{68.16} & \textbf{92.77} & \textbf{95.24} & \textbf{73.71} & \textbf{90.37} & \textbf{88.95 }&\textbf{ 75.81} & \textbf{89.59} & \textbf{69.20} & \textbf{79.53 }& \textbf{52.27} \\
\hline

\end{tabular}
\caption{Mean dice scores for the proposed models in all phase 1 tasks. All experiments were run as five-fold cross-validation. The models that we used for generating our test set submission are highlighted in bold. The dice scores of the test sets are shown at the bottom of the table. Test dice scores in bold denote that at the time of manuscript submission these scores were the highest in the online leaderboard of the challenge (\MYhref{https://decathlon.grand-challenge.org/evaluation/results}{decathlon.grand-challenge.org/evaluation/results}).}
\label{tab:results}
}

\end{table}


\section{Discussion}
In this paper we present the nnU-Net segmentation framework for the medical domain that directly builds around the original U-Net architecture \cite{ronneberger2015u} and dynamically adapts itself to the specifics of any given dataset. 
Based on our hypothesis that non-architectural modifications can be much more powerful than some of the recently presented architectural modifications, the essence of this framework is a thorough design of adaptive preprocessing, training scheme and inference. All design choices required to adapt to a new segmentation task are done in a fully automatic manner with no manual interaction.
For each task the nnU-Net automatically runs a five-fold cross-validation for three different automatically configures U-Net models and the model (or ensemble) with the highest mean foreground dice score is chosen for final submission. 
In the context of the Medical Segmentation Decathlon we demonstrate that the nnU-Net performs competitively on the held-out test sets of 7 highly distinct medical datasets, achieving the highest mean dice scores for all classes of all tasks (except class 1 in the BrainTumour dataset) on the online leaderboard at the time of manuscript submission. We acknowledge that training three models and picking the best one for each dataset independently is not the cleanest solution. Given a larger time-scale, one could investigate proper heuristics to identify the best model for a given dataset prior to training. Our current tendency favors the U-Net Cascade (or the 3D U-Net if the cascade cannot be applied) with the sole (close) exceptions being the Prostate and Liver tasks. Additionally, the added benefit of many of our design choices, such as the use of Leaky ReLUs instead of regular ReLUs and the parameters of our data augmentation were not properly validated in the context of this challenge. Future work will therefore focus on systematically evaluating all design choices via ablation studies.

\bibliographystyle{IEEEtran}
\bibliography{bibliography}
\end{document}